\documentclass{article}
\usepackage{amsmath,amssymb}
\usepackage{times}
\usepackage{subfigure}
\usepackage{capt-of}
\usepackage{graphicx}
\usepackage{url}
\usepackage{booktabs}
\usepackage{macros}
\usepackage{floatrow}
\usepackage[colorlinks,citecolor=blue,urlcolor=blue]{hyperref}
\usepackage{titlesec}
\usepackage[accepted]{icml2015}
\usepackage{floatrow}

\titleformat{\section}[hang]
  {\bfseries}{\thesection.}{1em}{}
\icmltitlerunning{Scalable Bayesian Optimization Using Deep Neural Networks}

\begin{document}

\twocolumn[\icmltitle{Scalable Bayesian Optimization Using Deep Neural Networks}
\icmlauthor{Jasper Snoek$^{*}$}{jsnoek@seas.harvard.edu}

\icmlauthor{Oren Rippel$^{\dagger*}$}{rippel@math.mit.edu}

\icmlauthor{Kevin Swersky$^{\S}$}{kswersky@cs.toronto.edu}

\icmlauthor{Ryan Kiros$^{\S}$}{rkiros@cs.toronto.edu}

\icmlauthor{Nadathur Satish$^{\ddagger}$}{nadathur.rajagopalan.satish@intel.com}

\icmlauthor{Narayanan Sundaram$^{\ddagger}$}{narayanan.sundaram@intel.com}

\icmlauthor{Md. Mostofa Ali Patwary$^{\ddagger}$}{mostofa.ali.patwary@intel.com}

\icmlauthor{Prabhat$^{\star}$}{prabhat@lbl.gov}

\icmlauthor{Ryan P. Adams$^{*}$}{rpa@seas.harvard.edu}

\small{
\icmladdress{$^{*}$Harvard University, School of Engineering and Applied Sciences}\vspace{-9pt}%
\icmladdress{$^{\dagger}$Massachusetts Institute of Technology, Department of Mathematics}\vspace{-9pt}%
\icmladdress{$^{\S}$University of Toronto, Department of Computer Science}\vspace{-9pt}%
\icmladdress{$^{\ddagger}$Intel Labs, Parallel Computing Lab}\vspace{-9pt}%
\icmladdress{$^{\star}$NERSC, Lawrence Berkeley National Laboratory}\vspace{-9pt}%
}
\icmlkeywords{bayesian optimization, deep learning}

\vskip 0.3in
]

\begin{abstract}
Bayesian optimization is an effective methodology for the global optimization of functions with expensive evaluations. It relies on querying a distribution over functions defined by a relatively cheap surrogate model.  An accurate model for this distribution over functions is critical to the effectiveness of the approach, and is typically fit using Gaussian processes (GPs).  However, since GPs scale cubically with the number of observations, it has been challenging to handle objectives whose optimization requires many evaluations, and as such, massively parallelizing the optimization.

In this work, we explore the use of neural networks as an alternative to GPs to model distributions over functions.  We show that performing adaptive basis function regression with a neural network as the parametric form performs competitively with state-of-the-art GP-based approaches, but scales linearly with the number of data rather than cubically.  This allows us to achieve a previously intractable degree of parallelism, which we apply to large scale hyperparameter optimization, rapidly finding competitive models on benchmark object recognition tasks using convolutional networks, and image caption generation using neural language models.
\end{abstract}

\section{Introduction}
Recently, the field of machine learning has seen unprecedented growth due to a new wealth of data, increases in computational power, new algorithms, and a plethora of exciting new applications. As researchers tackle more ambitious problems, the models they use are also becoming more sophisticated. However, the growing complexity of machine learning models inevitably comes with the introduction of additional hyperparameters. These range from design decisions such as the shape of a neural network architecture, to optimization parameters such as learning rates, to regularization hyperparameters such as weight decay. Proper setting of these hyperparameters is critical for performance on difficult problems.

There are many methods for optimizing over hyperparameter settings, ranging from simplistic procedures like grid or random search~\cite{bergstra-bengio-2012a}, to more sophisticated model-based approaches using random forests~\cite{hutter-2011a} or Gaussian processes~\cite{snoek-etal-2012b}. Bayesian optimization is a natural framework for model-based global optimization of noisy, expensive black-box functions. It offers a principled approach to modeling uncertainty, which allows exploration and exploitation  to be naturally balanced during the search. Perhaps the most commonly used model for Bayesian optimization is the Gaussian process (GP) due to its simplicity and flexibility in terms of conditioning and inference.

However, a major drawback of GP-based Bayesian optimization is that inference time grows cubically in the number of observations, as it necessitates the inversion of a dense covariance matrix. For problems with a very small number of hyperparameters, this has not been an issue, as the minimum is often discovered before the cubic scaling renders further evaluations prohibitive. As the complexity of machine learning models grows, however, the size of the search space grows as well, along with the number of hyperparameter configurations that need to be evaluated before a solution of sufficient quality is found. Fortunately, as models have grown in complexity, computation has become significantly more accessible and it is now possible to train many models in parallel. A natural solution to the hyperparameter search problem is to therefore combine large-scale parallelism with a scalable Bayesian optimization method. The cubic scaling of the GP, however, has made it infeasible to pursue this approach.

The goal of this work is to develop a method for scaling Bayesian optimization, while still maintaining its desirable flexibility and characterization of uncertainty. To that end, we propose the use of neural networks to learn an adaptive set of basis functions for Bayesian linear regression. We refer to this approach as Deep Networks for Global Optimization (DNGO). Unlike a standard Gaussian process, DNGO scales linearly with the number of function evaluations---which, in the case of hyperparameter optimization, corresponds to the number of models trained---and is amenable to stochastic gradient training. Although it may seem that we are merely moving the problem of setting the hyperparameters of the model being tuned to setting them for the tuner itself, we show that for a suitable set of design choices it is possible to create a robust, scalable, and effective Bayesian optimization system that generalizes across many global optimization problems.

We demonstrate the effectiveness of DNGO on a number of difficult problems, including benchmark problems for Bayesian optimization, convolutional neural networks for object recognition, and multi-modal neural language models for image caption generation. We find hyperparameter settings that achieve competitive with state-of-the-art results of $6.37\%$ and $27.4\%$ on CIFAR-10 and CIFAR-100 respectively, and BLEU scores of $25.1$ and $26.7$ on the Microsoft COCO 2014 dataset using a single model and a 3-model ensemble.

\section{Background and Related Work}

\subsection{Bayesian Optimization}
Bayesian optimization is a well-established strategy for the global optimization of noisy, expensive black-box functions~\cite{Mockus1978}.  For an in-depth review, see \citet{lizotte-thesis}, \citet{Brochu2010} and \citet{osborne-2009a}.  Bayesian optimization relies on the construction of a probabilistic model that defines a distribution over objective functions from the input space to the objective of interest.  Conditioned on a prior over the functional form and a set of $N$ observations of input-target pairs $\mathcal{D}={\{(\brmx_n, y_n)\}_{n=1}^{N}}$, the relatively cheap posterior over functions is then queried to reason about where to seek the optimum of the expensive function of interest. The promise of a new experiment is quantified using an \emph{acquisition function}, which, applied to the posterior mean and variance, expresses a trade-off between exploration and exploitation.  Bayesian optimization proceeds by performing a proxy optimization over this acquisition function in order to determine the next input to evaluate.

Recent innovation has resulted in significant progress in Bayesian optimization, including elegant theoretical results~\cite{Srinivas2010,Bull2011,defreitas-etal-2013a}, multitask and transfer optimization~\cite{krause-ong-2011,swersky-etal-2013a,remi2013collab} and the application to diverse tasks such as sensor set selection~\cite{garnett-etal-2010}, the tuning of adaptive Monte Carlo~\cite{mahendran-2012a} and robotic gait control~\cite{calandra-etal-2014a}.

Typically, GPs have been used to construct the distribution over functions used in Bayesian optimization, due to their flexibility, well-calibrated uncertainty, and analytic properties~\cite{Jones2001,osborne-2009a}.  Recent work has sought to improve the performance of the GP approach through accommodating higher dimensional problems~\cite{wang-etal-2013,djolonga13high}, input non-stationarities~\cite{snoek-etal-2014a} and initialization through meta-learning~\cite{feurer-2015}.  Random forests, which scale linearly with the data, have also been used successfully for algorithm configuration by \citet{hutter-2011a} with empirical estimates of model uncertainty.

More specifically, Bayesian optimization seeks to solve the minimization problem
\begin{align}
	\brmx^\star &= \argmin_{\brmx\in\mcX} f(\brmx)\,,
\end{align}
where we take~$\mcX$ to be a compact subset of~$\reals^K$.  In our work, we build upon the standard GP-based approach of~\citet{Jones2001} which uses a GP surrogate and the \emph{expected improvement} acquisition function~\cite{Mockus1978}.
For the surrogate model hyperparameters $\bTheta$, let~${\sigma^2(\brmx; \bTheta) = \Sigma(\brmx,\brmx; \bTheta)}$ be the marginal predictive variance of the probabilistic model, $\mu(\brmx; \mathcal{D}, \bTheta)$ be the predictive mean, and define
\begin{align}
\gamma(\brmx) &= \frac{f(\brmx_\text{best}) - \mu(\brmx; \mathcal{D}, \bTheta)}{\sigma(\brmx; \mathcal{D},\bTheta)}\;,
\end{align}
where $f(\brmx_\text{best})$ is the lowest observed value. The expected improvement criterion is defined as
\begin{align}
& a_{\text{EI}}(\brmx; \mathcal{D},\bTheta) =  \label{eqn:ei}\\
&\qquad \sigma(\brmx; \mathcal{D},\bTheta) \left[\gamma(\brmx) \Phi(\gamma(\brmx)) \right. \left. +\, \mathcal{N}(\gamma(\brmx); 0,1)\right] \;.\nonumber
\end{align}

Here $\Phi(\cdot)$ is the cumulative distribution function of a standard normal, and~$\mathcal{N}(\cdot; 0,1)$ is the density of a standard normal. Note that numerous alternate acquisition functions and combinations thereof have been proposed~\cite{kushner-1964a,Srinivas2010,hoffman-etal-2011}, which could be used without affecting the analytic properties of our approach.

\subsection{Bayesian Neural Networks}
The application of Bayesian methods to neural networks has a rich history in machine learning~\cite{mackay1992practical,hinton-camp-93,buntine1991bayesian,neal1995bayesian,de2003bayesian}. The goal of Bayesian neural networks is to uncover the full posterior distribution over the network weights in order to capture uncertainty, to act as a regularizer, and to provide a framework for model comparison. The full posterior is, however, intractable for most forms of neural networks, necessitating expensive approximate inference or Markov chain Monte Carlo simulation. More recently, full or approximate Bayesian inference has been considered for small pieces of the overall architecture. For example, in similar spirit to this work, \citet{lazaro2010marginalized,hinton2008using} and \citet{calandra2014manifold} considered inference over just the last layer of a neural network. Alternatively, variational approaches are developed in~\citet{kingma2013auto,rezende2014stochastic} and \citet{mnih2014neural}, where a neural network is used in a variational approximation to the posterior distribution over the latent variables of a directed generative neural network.

\section{Adaptive Basis Regression with Deep Neural Networks}

A key limitation of GP-based Bayesian optimization is that the computational cost of the technique scales cubically in the number of observations, limiting the applicability of the approach to objectives that require a relatively small number of observations to optimize.  In this work, we aim to replace the GP traditionally used in Bayesian optimization with a model that scales in a less dramatic fashion, but retains most of the GP's desirable properties such as flexibility and well-calibrated uncertainty.  Bayesian neural networks are a natural consideration, not least because of the theoretical relationship between Gaussian processes and infinite Bayesian neural networks~\cite{neal1995bayesian,williams-96}. However, deploying these at a large scale is very computationally expensive.

As such, we take a pragmatic approach and add a Bayesian linear regressor to the last hidden layer of a deep neural network, marginalizing only the output weights of the net while using a point estimate for the remaining parameters. This results in \emph{adaptive basis regression}, a well-established statistical technique which scales linearly in the number of observations, and cubically in the basis function dimensionality.  This allows us to explicitly trade off evaluation time and model capacity.  As such, we form the basis using the very flexible and powerful non-linear functions defined by the neural network.

First of all, without loss of generality and assuming compact support for each input dimension, we scale the input space to the unit hypercube. We denote by~$\bphi(\cdot)=[\phi_1(\cdot),\ldots,\phi_D(\cdot)]^{\trans}$ the vector of outputs from the last hidden layer of the network, trained on inputs and targets~${\mathcal{D}:=\left\{(\brmx_n, y_n)\right\}_{n=1}^N\subset\mathbb{R}^K\times \mathbb R}$. We take these to be our set of basis functions. In addition, define~$\bPhi$ to be the design matrix arising from the data and this basis, where~${\Phi_{nd}=\phi_d(\brmx_n)}$ is the output design matrix, and $\brmy$ the stacked target vector.

These basis functions are parameterized via the weights and biases of the deep neural network, and these parameters are trained via backpropagation and stochastic gradient descent with momentum.  In this training phase, a linear output layer is also fit.  This procedure can be viewed as a maximum \emph{a posteriori} (MAP) estimate of all parameters in the network.  Once this ``basis function neural network'' has been trained, we replace the MAP-parameterized output layer with a Bayesian linear regressor that captures uncertainty in the weights.  See Section \ref{marginal} for a more elaborate explanation of this choice.

The predictive mean $\mu(\brmx; \bTheta)$ and variance $\sigma^2(\brmx; \bTheta)$ of the model are then given by~\citep[see][]{bishop-2006}
\begin{align}
\mu(\brmx; \mathcal{D}, \bTheta) &= \brmm^{\trans}\bphi(\brmx) + \eta(\brmx) \label{eq:postpredmean}\;,\\
\sigma^2(\brmx; \mathcal{D},\bTheta) &= \bphi(\brmx)^{\trans}\brmK^{-1}\bphi(\brmx) + \frac{1}{\beta} \label{eq:postpredvar}
\end{align}
where
\begin{align}
\brmm&=\beta \brmK^{-1}\bPhi^T\brmyhat\in\mathbb{R}^{D} \\
\brmK &= \beta\bPhi^T\bPhi + \bI\alpha\in\mathbb{R}^{D\times D}.
\end{align}

Here, $\eta(\brmx)$ is a prior mean function which is described in Section~\ref{quadratic}, and $\brmyhat = \brmy-\eta(\brmx)$. In addition, ${\alpha, \beta \in \bTheta}$ are regression model hyperparameters.  We integrate out~$\alpha$ and~$\beta$ using slice sampling~\cite{Neal00slicesampling} according to the methodology of~\citet{snoek-etal-2012b} over the marginal likelihood, which is given by
\begin{align}
\log p(\brmy&\given\brmX,\alpha,\beta) = \frac{D}{2}\log\alpha + \frac{N}{2}\log\beta - \frac{N}{2}\log(2\pi) \nonumber\\
&- \frac{\beta}{2}||\brmyhat-\bPhi\brmm||^2 - \frac{\alpha}{2}\brmm^{\trans}\brmm - \frac{1}{2}\log|\brmK|\;.
\label{eqn:marginallikelihood}
\end{align}

\begin{figure}[t]
  \includegraphics[width=\textwidth]{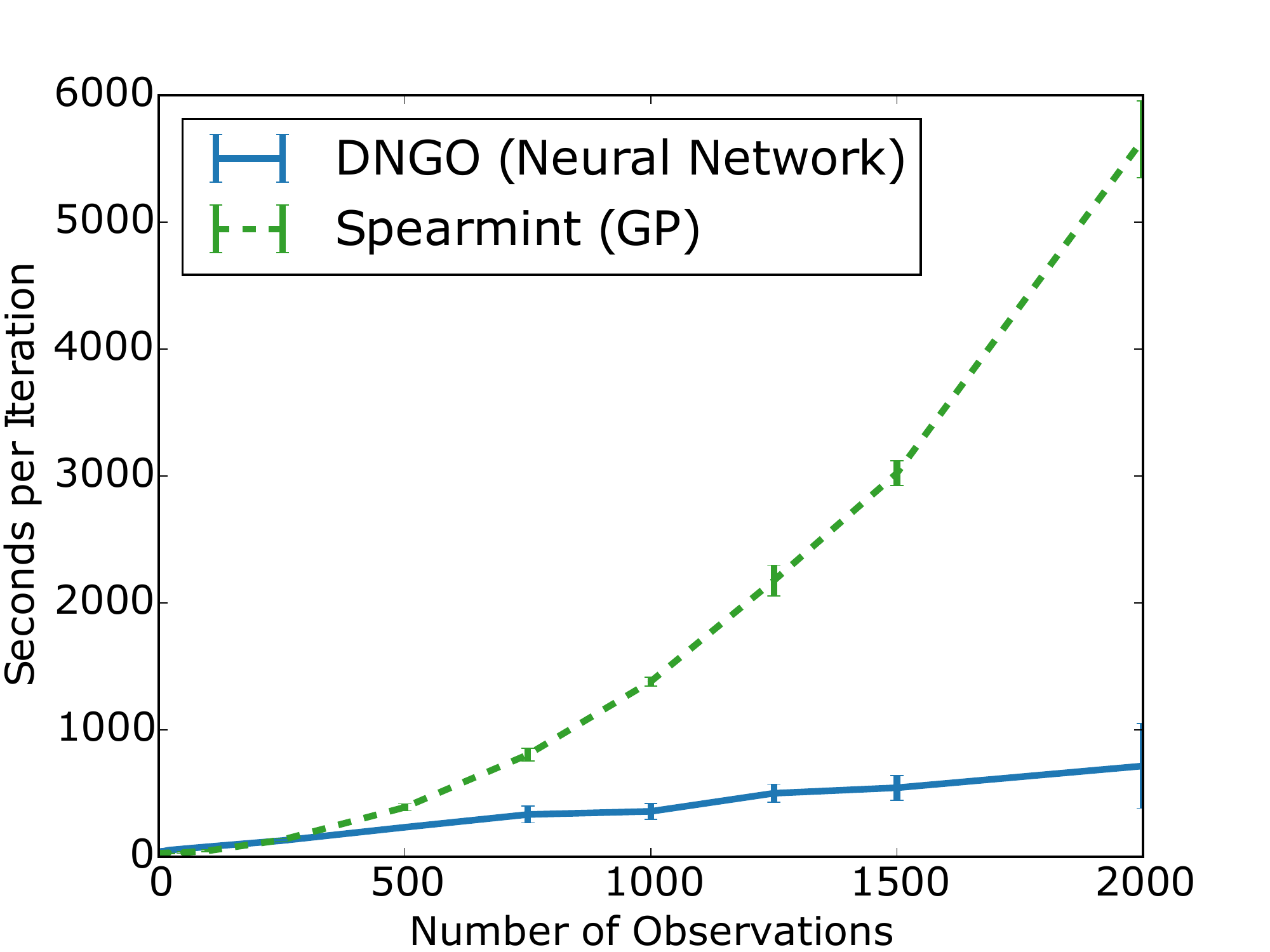}%
\caption{The time per suggested experiment for our method compared to the state-of-the-art GP based approach from~\citet{snoek-etal-2014a} on the six dimensional Hartmann function.  We ran each algorithm on the same 32 core system with 80GB of RAM five times and plot the mean and standard deviation.}
\label{fig:timings}
\end{figure}

It is clear that the computational bottleneck of this procedure is the inversion of $\brmK$. However, note that the size of this matrix grows with the output dimensionality $D$, rather than the number of observations $N$ as in the GP case.  This allows us to scale to significantly more observations than with the GP as demonstrated in Figure~\ref{fig:timings}.

\subsection{Model details}

\subsubsection{Network architecture}
\label{network}
\begin{figure*}[t]
\centering%
\subfigure[\small TanH Units]{
  \includegraphics[width=0.28\linewidth,trim=1cm 1cm 1cm 1cm,clip]{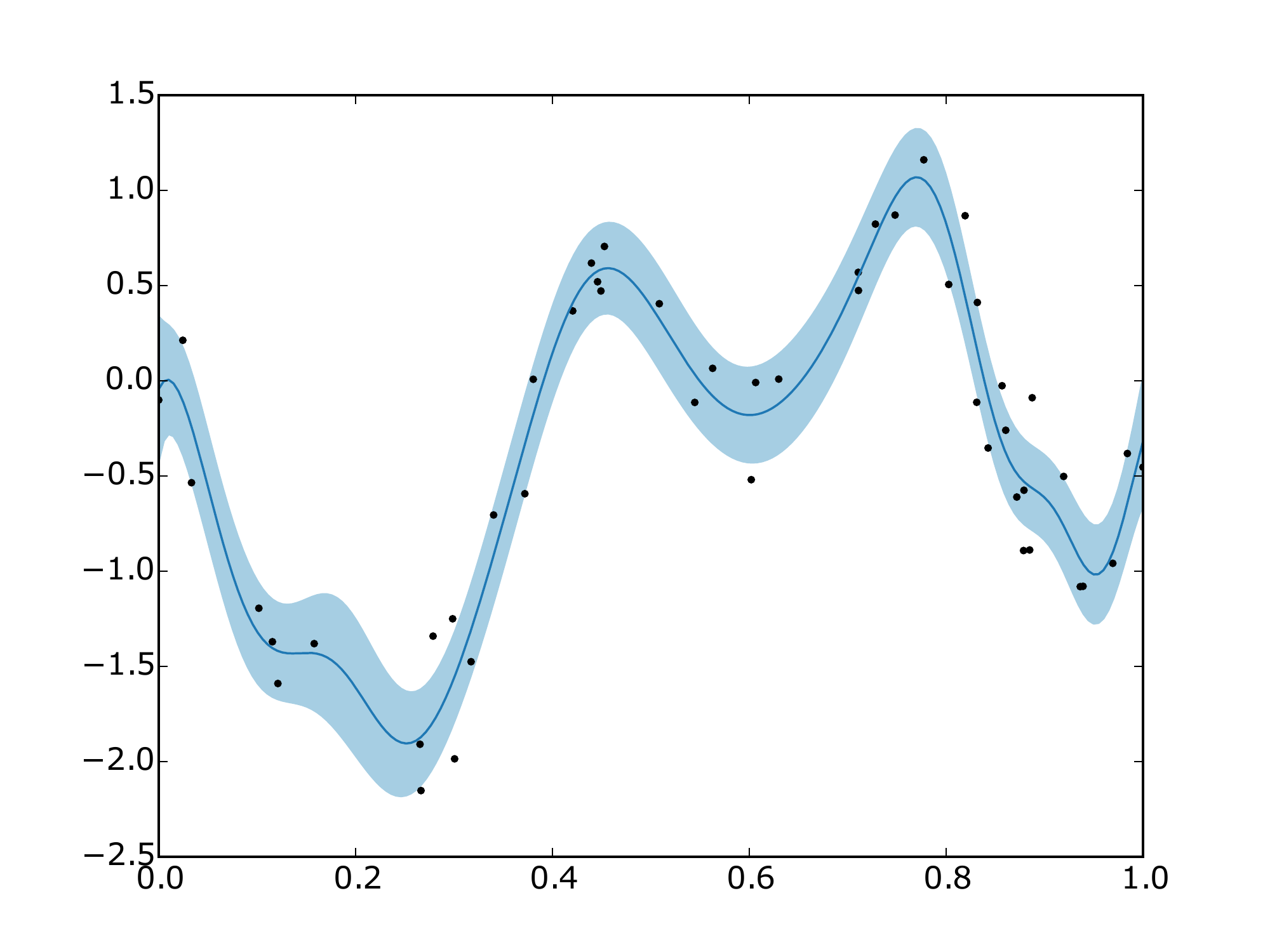}%
  \label{fig:tanh}}
\subfigure[\small ReLU + TanH Units]{
  \includegraphics[width=0.28\linewidth,trim=1cm 1cm 1cm 1cm,clip]{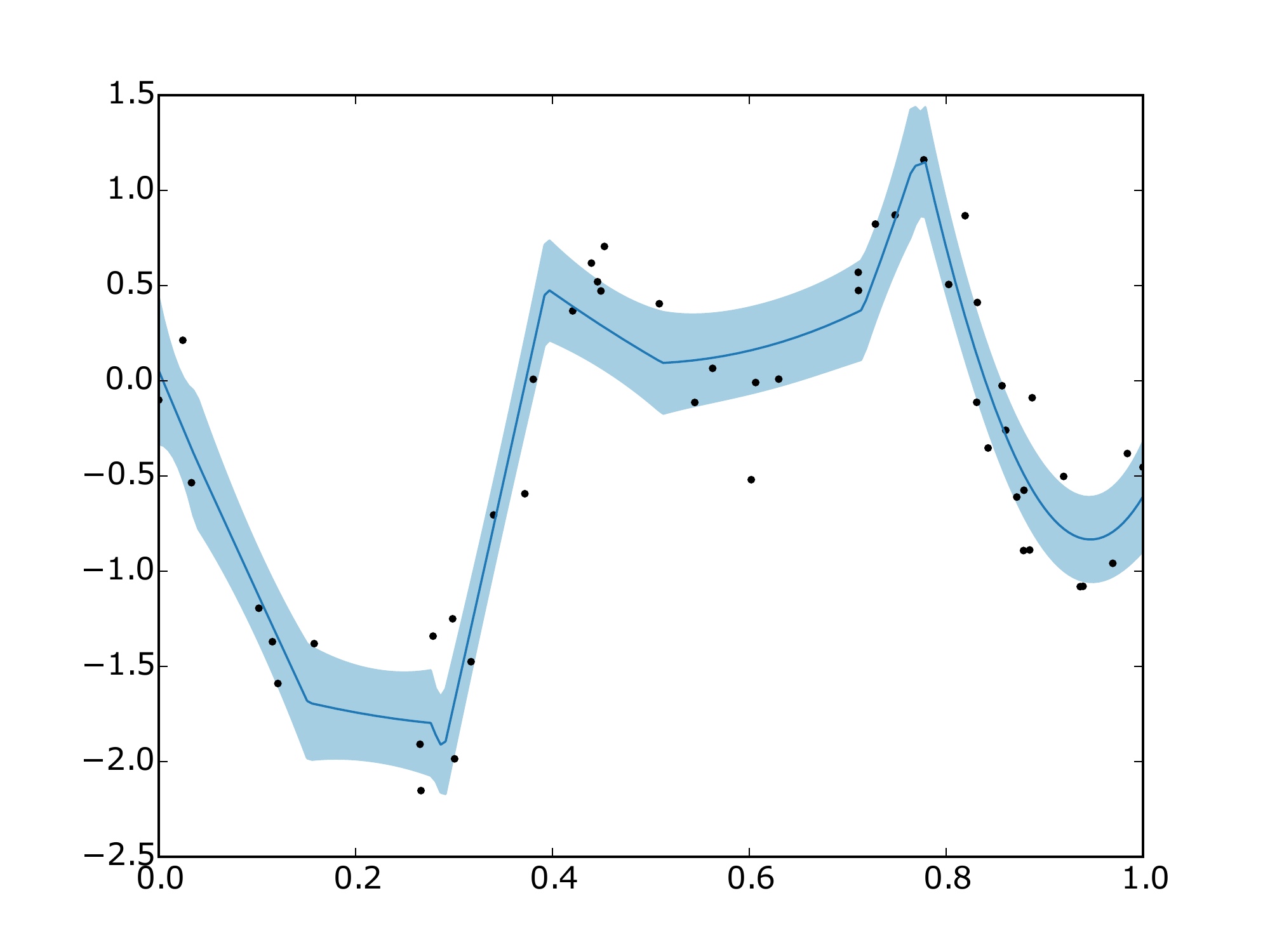}%
  \label{fig:relutanh}}
\subfigure[\small ReLU Units]{
  \includegraphics[width=0.28\linewidth,trim=1cm 1cm 1cm 1cm,clip]{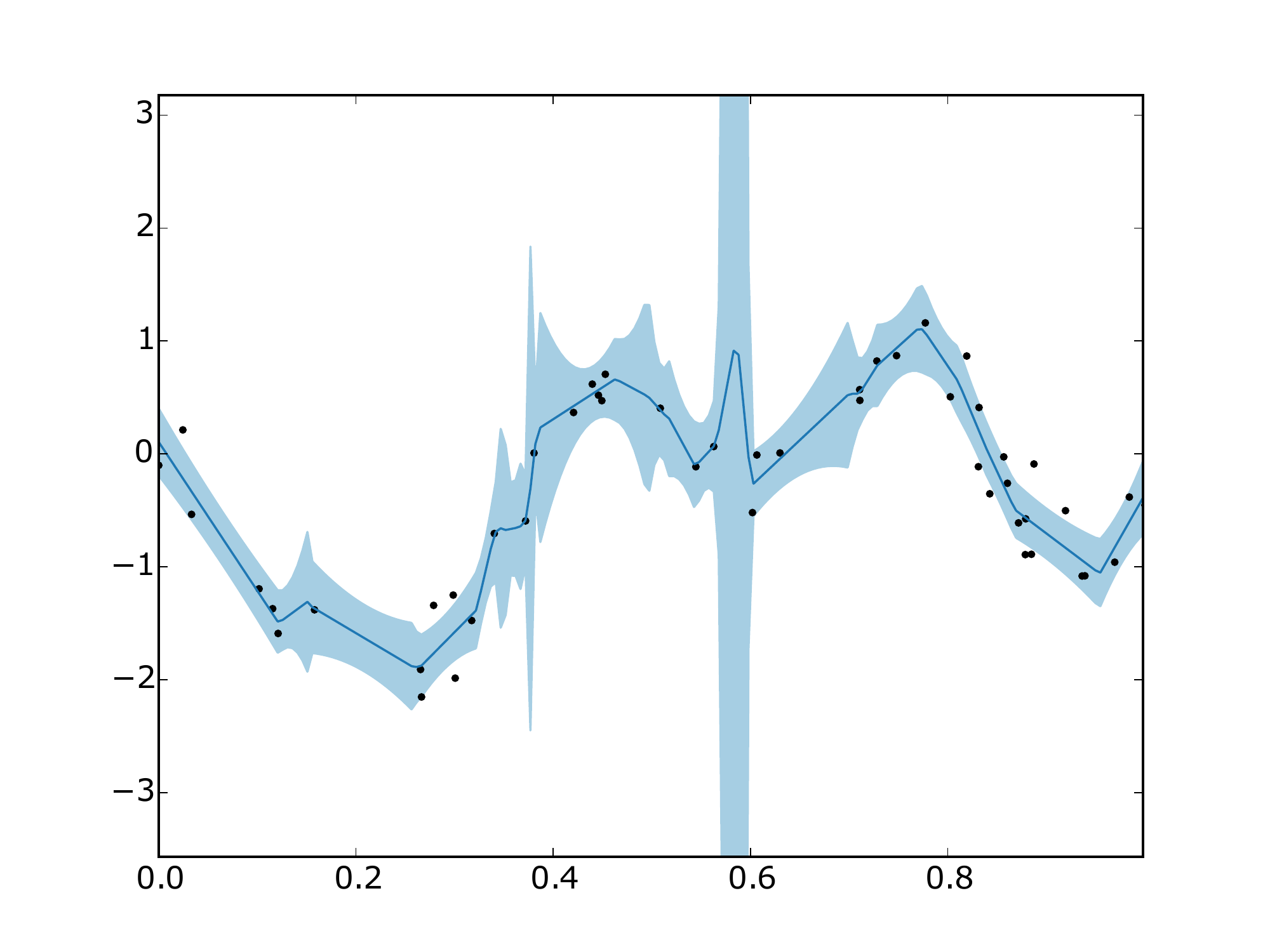}%
  \label{fig:relu}}
\hfill\subfigure[]{\centering
\includegraphics[width=0.06\linewidth,trim=0 0 0 0,clip]{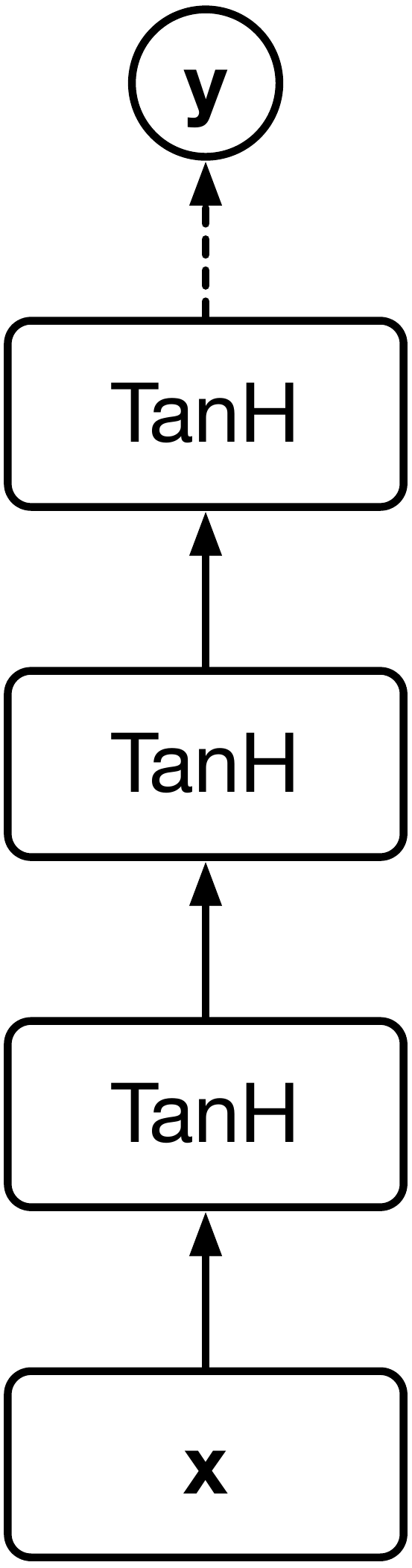}%
\label{fig:architecture}}
\vspace{-9pt}
\caption{A comparison of the predictive mean and uncertainty learned by the model when using \ref{fig:tanh}~only tanh, \ref{fig:relu}~only rectified linear (ReLU) activation functions or \ref{fig:relutanh}~ReLU's but a tanh on the last hidden layer.  The shaded regions correspond to standard deviation envelopes around the mean. The choice of activation function significantly modifies the basis functions learned by the model.  Although the ReLU, which is the standard for deep neural networks, is highly flexible, we found that its unbounded activation can lead to extremely large uncertainty estimates. Subfigure~\ref{fig:architecture} illustrates the overall architecture of the DNGO model. Dashed lines correspond to weights that are marginalized.
}
\label{fig:activation_functions}

\end{figure*}

A natural concern with the use of deep networks is that they often require significant effort to tune and tailor to specific problems. One can consider adjusting the architecture and tuning the hyperparameters of the neural network as itself a difficult hyperparameter optimization problem.  An additional challenge is that we aim to create an approach that generalizes across optimization problems.  We found that design decisions such as the type of activation function used significantly altered the performance of the Bayesian optimization routine.  For example, in Figure~\ref{fig:activation_functions} we see that the commonly used rectified linear (ReLU) function can lead to very poor estimates of uncertainty, which causes the Bayesian optimization routine to explore excessively.  Since the bounded tanh function results in smooth functions with realistic variance, we use this nonlinearity in this work; however, if the smoothness assumption needs to be relaxed, a combination of rectified linear functions with a tanh function only on the last layer can also be used in order to bound the basis.

In order to tune any remaining hyperparameters, such as the width of the hidden layers and the amount of regularization, we used GP-based Bayesian optimization.  For each of one to four layers we ran Bayesian optimization using the Spearmint~\cite{snoek-etal-2014a} package to minimize the average relative loss on a series of benchmark global optimization problems.  We tuned a global learning rate, momentum, layer sizes, $\ell_2$ normalization penalties for each set of weights and dropout rates~\cite{hinton2012improving} for each layer.  Interestingly, the optimal configuration featured no dropout and very modest $\ell_2$ normalization.  We suspect that dropout, despite having an approximate correction term, causes noise in the predicted mean resulting in a loss of precision.  The optimizer instead preferred to restrict capacity via a small number of hidden units. Namely, the optimal architecture is a deep and narrow network with 3 hidden layers and approximately 50 hidden units per layer.  We use the same architecture throughout our empirical evaluation, and this architecture is illustrated in Figure~\ref{fig:architecture}.

\subsubsection{Marginal likelihood vs MAP estimate}
\label{marginal}
The standard empirical Bayesian approach to adaptive basis regression is to maximize the marginal likelihood with respect to the parameters of the basis (see Equation~\ref{eqn:marginallikelihood}), thus taking the model uncertainty into account.  However, in the context of our method, this requires evaluating the gradient of the marginal likelihood, which requires inverting a~${D\times D}$ matrix on each update of stochastic gradient descent.  As this makes the optimization of the net significantly slower, we take a pragmatic approach and optimize the basis using a point estimate and apply the Bayesian linear regression layer \emph{post-hoc}.  We found that both approaches gave qualitatively and empirically similar results, and as such we in practice employ the more efficient one.

\subsubsection{Quadratic Prior}
\label{quadratic}
One of the advantages of Bayesian optimization is that it provides natural avenues for incorporating prior information about the objective function and search space. For example, when choosing the boundaries of the search space, a typical assumption has been that the optimal solution lies somewhere in the interior of the input space. However, by the curse of dimensionality, most of the volume of the space lies very close to its boundaries. Therefore, we select a mean function $\eta(\brmx)$ (see Equation~\ref{eq:postpredmean}) to reflect our subjective prior beliefs that the function is coarsely approximated by a convex quadratic function centered in the bounded search region, i.e.,
\begin{align}
\eta(\brmx) = \lambda + (\brmx - \brmc)^T \bLambda (\brmx - \brmc)
\end{align}
where $\brmc$ is the center of the quadratic,~$\lambda$ is an offset and~$\bLambda$ a diagonal scaling matrix.  We place a Gaussian prior with mean $0.5$ (the center of the unit hypercube) on $\brmc$, horseshoe~\cite{carvalho-2009a} priors on the diagonal elements $\Lambda_{kk}\; \forall k\in\{1,\ldots, K\}$ and integrate out $b$, $\lambda$ and $\brmc$ using slice sampling over the marginal likelihood.

The horseshoe is a so-called \emph{one-group} prior for inducing sparsity and is a somewhat unusual choice for the weights of a regression model.  Here we choose it because it 1) has support only on the positive reals, leading to convex functions, and 2) it has a large spike at zero with a heavy tail, resulting in strong shrinkage for small values while preserving large ones.  This last effect is important for handling model misspecification as it allows the quadratic effect to disappear and become a simple offset if necessary.

\begin{table*}[t]
  \centering
  \ffigbox{
  \begin{tabular}[b]{lrrrr|r}
    \toprule
    Experiment & \# Evals & \multicolumn{1}{c}{SMAC} & \multicolumn{1}{c}{TPE} & \multicolumn{1}{c|}{Spearmint} & \multicolumn{1}{c}{DNGO} \\%
    \cmidrule(r){1-6}%
    Branin (0.398) & 200 & $0.655\pm 0.27$ & $0.526\pm 0.13$ & $\bf 0.398\pm 0.00$ & $\bf 0.398\pm 0.00$\\
    Hartmann6 (-3.322) & 200 & $-2.977\pm 0.11$ & $-2.823\pm 0.18$ & $\bf -3.3166\pm 0.02$ & $-3.319 \pm 0.00$\\
    Logistic Regression & 100 & $8.6\pm 0.9$ & $8.2\pm 0.6$ & $\bf 6.88 \pm 0.0$ & $\bf 6.89 \pm 0.04$\\
    LDA (On grid) & 50 & $1269.6\pm 2.9$ & $1271.5\pm 3.5$ & $\bf 1266.2 \pm 0.1$ & $\bf 1266.2 \pm 0.0$ \\
    SVM (On grid) & 100 & $\bf 24.1\pm 0.1$ & $24.2\pm 0.0$ & $\bf 24.1 \pm 0.1$ & $\bf 24.1 \pm 0.1$\\
   \bottomrule%
\end{tabular}}{%
  \captionsetup[table]{position=bottom}
  \captionof{table}{Evaluation of DNGO on global optimization benchmark problems versus scalable (TPE, SMAC) and non-scalable (Spearmint) Bayesian optimization methods. All problems are minimization problems. For each problem, each method was run $10$ times to produce error bars.
  }
  \label{tab:result_comparison}
  }
\end{table*}

\subsection{Incorporating input space constraints}
Many problems of interest have complex, possibly unknown bounds, or exhibit undefined behavior in some regions of the input space.  These regions can be characterized as \emph{constraints} on the search space.  Recent work~\cite{gelbart-etal-2014,snoek-2013a,gramacy2010} has developed approaches for modeling unknown constraints in GP-based Bayesian optimization by learning a constraint classifier and then discounting expected improvement by the probability of constraint violation.

More specifically, define ${c_n \in \{0,1\}}$ to be a binary indicator of the validity of input $\brmx_n$. Also, denote the sets of valid and invalid inputs as ${\mathcal{V}=\left\{(\brmx_n, y_n)\;|\; c_n=1\right\}}$ and ${\mathcal{I}=\left\{(\brmx_n, y_n)\;|\; c_n=0\right\}}$, respectively. Note that ${\mathcal{D}:=\mathcal{V}\cup\mathcal{I}}$. Lastly, let $\bPsi$ be the collection of constraint hyperparameters. The modified expected improvement function can be written as
\begin{align}
  a_{\mathrm{CEI}}(\brmx; \mathcal{D},\bTheta,\bPsi) = \;a_{\mathrm{EI}}(\brmx; \mathcal{V},\bTheta)\mathbb{P}\left[c=1\given \brmx,\mathcal{D},\bPsi\right]\;.\nonumber
\end{align}

In this work, to model the constraint surface, we similarly replace the Gaussian process with the adaptive basis model, integrating out the output layer weights:
\begin{align}
\begin{split}
  \mathbb{P}[ c&=1\given \brmx,\mathcal{D},\bPsi] = \\
  &\int_{\brmw} \mathbb{P}\left[c=1\given \brmx, \mathcal{D}, \brmw,\bPsi\right]\mathrm{P}(\brmw;\bPsi) \mathrm{d}\brmw\;.
\end{split}
\end{align}

In this case, we use a Laplace approximation to the posterior.  For noisy constraints we perform Bayesian logistic regression, using a logistic likelihood function for $\mathbb{P}\left[c=1\given \brmx, \mathcal{D}, \brmw,\bPsi\right]$.  For noiseless constraints, we replace the logistic function with a step function.

\subsection{Parallel Bayesian Optimization}
Obtaining a closed form expression for the joint acquisition function across multiple inputs is intractable in general~\cite{Ginsbourger2010a}.  However, a successful Monte Carlo strategy for parallelizing Bayesian optimization was developed in~\citet{snoek-etal-2012b}. The idea is to marginalize over the possible outcomes of currently running experiments when making a decision about a new experiment to run. Following this strategy, we use the posterior predictive distribution given by Equations \ref{eq:postpredmean} and \ref{eq:postpredvar} to generate a set of fantasy outcomes for each running experiment which we then use to augment the existing dataset. By averaging over sets of fantasies, we can perform approximate marginalization when computing EI for a candidate point. We note that this same idea works with the constraint network, where instead of computing marginalized EI, we would compute the marginalized probability of violating a constraint.

To that end, given currently running jobs with inputs $\{\brmx_j\}_{j=1}^J$, the marginalized acquisition function $a_{\mathrm{MCEI}}(\cdot; \mathcal{D}, \bTheta, \Psi)$ is given by
\begin{align}
\begin{split}
  a_{\mathrm{MCEI}} &(\brmx; \mathcal{D}, \{\brmx_j\}_{j=1}^J, \bTheta, \Psi)= \\
  &\int a_{\mathrm{CEI}}(\brmx; \mathcal{D}\cup\{(\brmx_j, y_j)\}_{j=1}^J, \bTheta, \Psi) \\
  &\times\mathbb{P}\left[\{c_j,y_j\}_{j=1}^J \given \mathcal{D}, \{\brmx\}_{j=1}^J\right]\mathrm{d}y_1...\mathrm{d}y_n \mathrm{d}c_1...\mathrm{d}c_n\;. \nonumber
\end{split}
\end{align}

When this strategy is applied to a GP, the cost of computing EI for a candidate point becomes cubic in the size of the augmented dataset. This restricts both the number of running experiments that can be tolerated, as well as the number of fantasy sets used for marginalization. With DNGO it is possible to scale both of these up to accommodate a much higher degree of parallelism. %

Finally, following the approach of~\citet{snoek-etal-2012b} we integrate out the hyperparameters of the model to obtain our final integrated acquisition function.  For each iteration of the optimization routine we pick the next input, $\brmx^{\mathrm{*}}$, to evaluate according to
\begin{align}
\brmx^{\mathrm{*}} &= \argmax_{\brmx} a_{\mathrm{MCEI}}(\brmx; \mathcal{D}, \{\brmx_j\}_{j=1}^J)\;,
\end{align}
where
\vspace{-9pt}
\begin{align}
\begin{split}
a_{\mathrm{MCEI}}&(\brmx; \mathcal{D}, \{\brmx_j\}_{j=1}^J) = \\
&\int a_{\mathrm{MCEI}}(\brmx; \mathcal{D}, \{\brmx_j\}_{j=1}^J, \bTheta, \bPsi)\; \mathrm{d}\bTheta \mathrm{d}\bPsi.
\end{split}
\end{align}

\begin{figure*}[t]
\centering%
\subfigure[``A person riding a wave in the ocean.'']{
  \includegraphics[height=1.5in,trim=1cm 1cm 1cm 1cm,clip]{success1.jpg}%
  \label{fig:success1}
}\hfill
\subfigure[``A bird sitting on top of a field.'']{
  \includegraphics[height=1.5in,trim=1cm 1cm 1cm 1cm,clip]{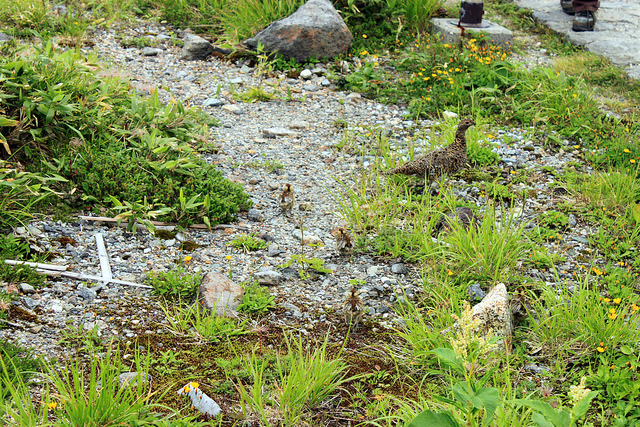}%
  \label{fig:success2}
}\hfill
\subfigure[``A horse is riding a horse.'']{
  \includegraphics[height=1.5in,trim=1cm 1cm 1cm 1cm,clip]{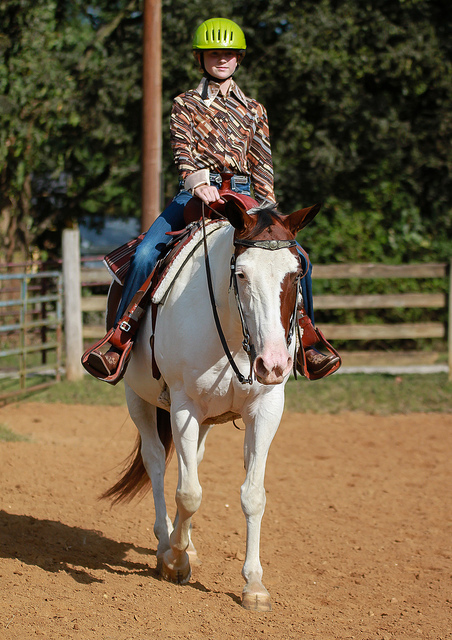}%
  \label{fig:failure}
}
\vspace{-9pt}
\caption{Sample test images and generated captions from the best LBL model on the COCO 2014 dataset. The first two captions sensibly describe the contents of their respective images, while the third is offensively inaccurate.}
\label{fig:captions}
\end{figure*}

\section{Experiments}
\label{sec:empirical}
\subsection{HPOLib Benchmarks}
In the literature, there exist several other methods for model-based optimization. Among these, the most popular variants in machine learning are the random forest-based SMAC procedure~\cite{hutter-2011a} and the tree Parzen estimator (TPE)~\cite{BergstraJ2011}. These are faster to fit than a Gaussian process and scale more gracefully with large datasets, but this comes at the cost of a more heuristic treatment of uncertainty. By contrast, DNGO provides a balance between scalability and the Bayesian marginalization of model parameters and hyperparameters.

To demonstrate the effectiveness of our approach, we compare DNGO to these scalable model-based optimization variants, as well as the input-warped Gaussian process method of~\citet{snoek-etal-2014a} on the benchmark set of continuous problems from the HPOLib package~\cite{Eggensperger-etal-2013a}. As Table \ref{tab:result_comparison} shows, DNGO significantly outperforms SMAC and TPE, and is competitive with the Gaussian process approach. This shows that, despite vast improvements in scalability, DNGO retains the statistical efficiency of the Gaussian process method in terms of the number of evaluations required to find the minimum.

\begin{table}[t]
  \centering
  \begin{tabular}[b]{lr}
    \toprule
    {\bf Method} & {\bf Test BLEU}\\%
    \cmidrule(r){1-2}%
    Human Expert LBL & $24.3$\\
    Regularized LSTM & $24.3$\\
    Soft-Attention LSTM & $24.3$\\
    10 LSTM ensemble & $24.4$\\
    Hard-Attention LSTM & $25.0$\\
    \cmidrule(r){1-2}%
    Single LBL & $\mathbf{25.1}$\\
    2 LBL ensemble & $\mathbf{25.9}$\\
    3 LBL ensemble & $\mathbf{26.7}$\\
   \bottomrule%
\end{tabular}
  \captionsetup[table]{position=bottom}
  \captionof{table}{Image caption generation results using BLEU-4 on the Microsoft COCO 2014 test set.
  Regularized and ensembled LSTM results are reported in \citet{zaremba-etal-2015a}. The baseline LBL tuned by a human expert and the Soft and Hard Attention models are reported in~\citet{xu-etal-2015a}.  We see that ensembling our top models resulting from the optimization further improves results significantly.  We noticed that there were distinct multiple local optima in the hyperparameter space, which may explain the dramatic improvement from ensembling a small subset of models.}
  \label{tab:captionresults}
	\vspace{-13pt}
\end{table}

\subsection{Image Caption Generation}
In this experiment, we explore the effectiveness of DNGO on a practical and expensive problem where highly parallel evaluation is necessary to make progress in a reasonable amount of time.  We consider the task of image caption generation using multi-modal neural language models. Specifically, we optimize the hyperparameters of the log-bilinear model (LBL) from~\citet{kiros-etal-2014a} to maximize the BLEU score of a validation set from the recently released COCO dataset~\cite{lin2014microsoft}.  In our experiments, each evaluation of this model took an average of 26.6 hours.

We optimize learning parameters such as learning rate, momentum and batch size; regularization parameters like dropout and weight decay for word and image representations; and architectural parameters such as the context size, whether to use the additive or multiplicative version, the size of the word embeddings and the multi-modal representation size~\footnote{Details are provided in the supplementary material.}. The final parameter is the number of factors, which is only relevant for the multiplicative model. This adds an interesting challenge, since it is only relevant for half of the hyperparameter space. This gives a total of 11 hyperparameters. Even though this number seems small, this problem offers a number of challenges which render its optimization quite difficult. For example, in order to not lose any generality, we choose broad box constraints for the hyperparameters; this, however, renders most of the volume of the model space infeasible. In addition, quite a few of the hyperparameters are categorical, which introduces severe non-stationarities in the objective surface.

Nevertheless, one of the advantages of a scalable method is the ability to highly parallelize hyperparameter optimization. In this way, high quality settings can be found after only a few sequential steps. To test DNGO in this scenario, we optimize the log-bilinear model with up to 800 parallel evaluations.

Running between 300 and 800 experiments in parallel (determined by cluster availability), we proposed and evaluated approximately 2500 experiments---the equivalent of over 2700 CPU days---in less than one week. Using the BLEU-4 metric, we optimized the validation set performance and the best LBL model found by DNGO outperforms recently proposed models using LSTM recurrent neural networks~\cite{zaremba-etal-2015a,xu-etal-2015a} on the test set. This is remarkable, as the LBL is a relatively simple approach. Ensembling this top model with the second and third best (under the validation metric) LBL models resulted in a test-set BLEU score~\footnote{We have verified that our BLEU score evaluation is consistent across reported results. We used a beam search decoding for our test predictions with the LBL model.} of $26.7$, significantly outperforming the LSTM-based approaches. We noticed that there were distinct multiple local optima in the hyperparameter space, which may explain the dramatic improvement from ensembling a small number of models.  We show qualitative examples of generated captions on test images in Figure~\ref{fig:captions}.  Further figures showing the BLEU score as a function of the iteration of Bayesian optimization are provided in the supplementary.

\begin{table}[t]
  \centering
  \small{
  \begin{tabular}[b]{lrrr}
    \toprule
    {\bf Layer type} & \# {\bf Filters} & {\bf Window} & {\bf Stride}\\%
    \cmidrule(r){1-4}%
    Convolution & $96$ & $3\times 3$ & \\
    \cmidrule(r){1-4}%
    Convolution & $96$ & $3\times 3$ & \\
    \cmidrule(r){1-4}%
    Max pooling & & $3\times 3$ & 2 \\
    \cmidrule(r){1-4}%
    Convolution & $192$ & $3\times 3$ & \\
    \cmidrule(r){1-4}%
    Convolution & $192$ & $3\times 3$ & \\
    \cmidrule(r){1-4}%
    Convolution & $192$ & $3\times 3$ & \\
    \cmidrule(r){1-4}%
    Max pooling & & $3\times 3$ & 2 \\
    \cmidrule(r){1-4}%
    Convolution & $192$ & $3\times 3$ & \\
    \cmidrule(r){1-4}%
    Convolution & $192$ & $1\times 1$ & \\
    \cmidrule(r){1-4}%
    Convolution & $10/100$ & $1\times 1$ & \\
    \cmidrule(r){1-4}%
    Global averaging & & $6\times 6$ & \\
    \cmidrule(r){1-4}%
    Softmax & & & \\
   \bottomrule%
  \end{tabular}}
  \captionsetup[table]{position=bottom}
  \captionof{table}{Our convolutional neural network architecture. This choice was chosen to be maximally generic. Each convolution layer is followed by a ReLU nonlinearity.}
  \label{tab:architecture}
\end{table}

\subsection{Deep Convolutional Neural Networks}
Finally, we use DNGO on a pair of highly competitive deep learning visual object recognition benchmark problems.  We tune the hyperparameters of a deep convolutional neural network on the CIFAR-10 and CIFAR-100 datasets. Our approach is to establish a single, generic architecture, and specialize it to various tasks via individualized hyperparameter tuning. As such, for both datasets, we employed the same generic architecture inspired by the configuration proposed in \citet{DBLP:journals/corr/SpringenbergDBR14}, which was shown to attain strong classification results. This architecture is detailed in Table~\ref{tab:architecture}.

\begin{figure}[t]
  \includegraphics[width=\textwidth,trim=0 0 0 0,clip]{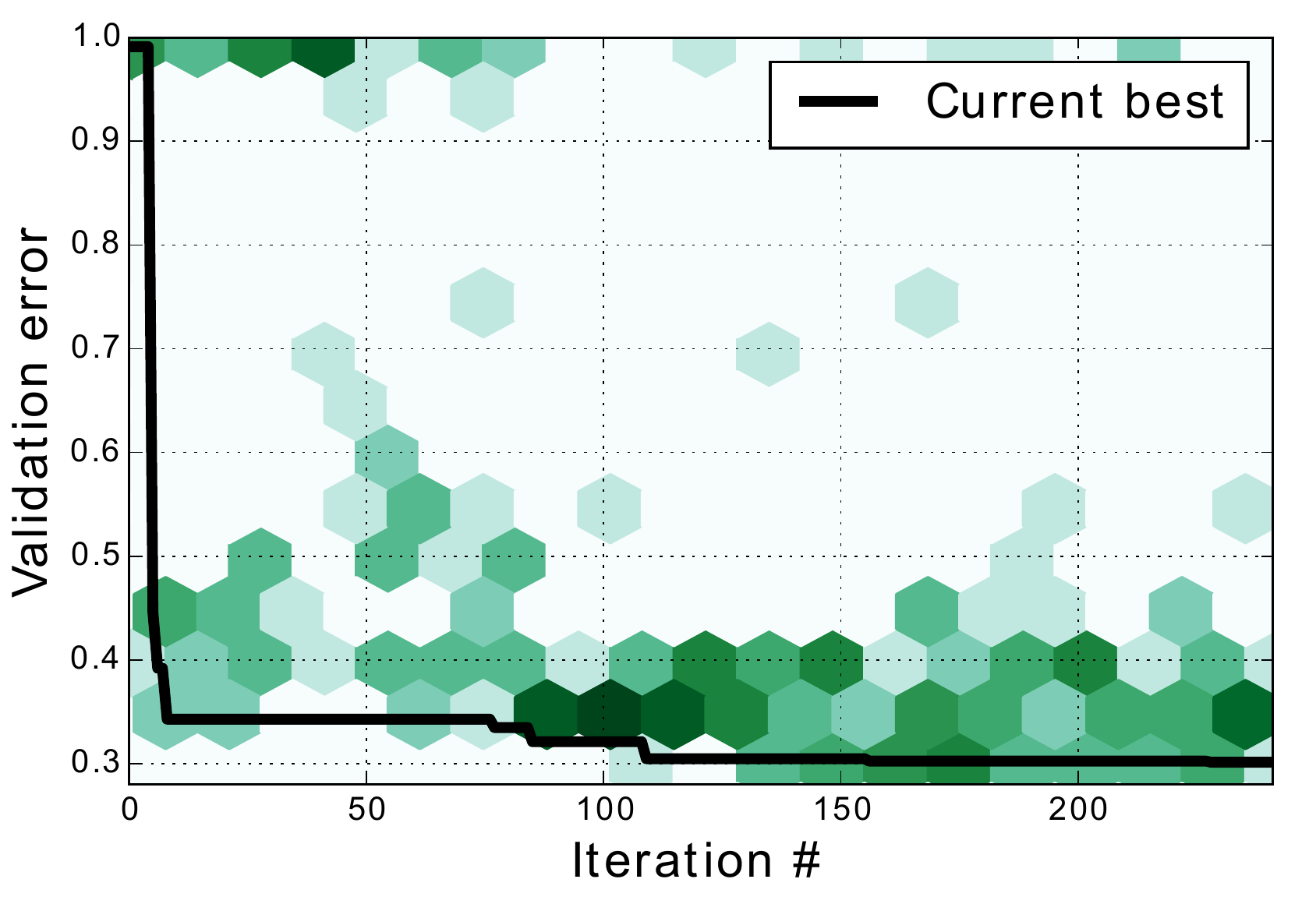}%
  \vspace{-9pt}
\caption{Validation errors on CIFAR-100 corresponding to different hyperparameter configurations as evaluated over time. These are represented as a planar histogram, where the shade of each bin indicates the total count within it. The current best validation error discovered is traced in black. This projection demonstrates the exploration-versus-exploitation paradigm of Bayesian optimization, in which the algorithm trades off visiting unexplored parts of the space, and focusing on parts which show promise.}
\label{fig:validation_errors}
\vspace{-9pt}
\end{figure}

For this architecture, we tuned the momentum, learning rate, $\ell_2$ weight decay coefficients, dropout rates, standard deviations of the random i.i.d.\ Gaussian weight initializations, and corruption bounds for various data augmentations: global perturbations of hue, saturation and value, random scalings, input pixel dropout and random horizontal reflections. We optimized these over a validation set of 10,000 examples drawn from the training set, running each network for 200 epochs. See Figure \ref{fig:validation_errors} for a visualization of the hyperparameter tuning procedure.

\begin{table}[t!]
  \centering
  \begin{tabular}[b]{lrr}
    \toprule
    {\bf Method} & {\bf CIFAR-10} & {\bf CIFAR-100}\\%
    \cmidrule(r){1-3}%
    Maxout & $9.38\%$ & $38.57\%$ \\
    DropConnect & $9.32\%$ & N/A \\
    Network in network & $8.81\%$ & $35.68\%$ \\
    Deeply supervised & $7.97\%$ & $34.57\%$ \\%
    ALL-CNN & $7.25\%$ & $33.71\%$ \\%
    \cmidrule(r){1-3}%
    Tuned CNN & $\bf 6.37\%$ & $\bf 27.4\%$ \\
   \bottomrule%
\end{tabular}
  \captionsetup[table]{position=bottom}
  \captionof{table}{We use our algorithm to optimize validation set error as a function of various hyperparameters of a convolutional neural network. We report the test errors of the models with the optimal hyperparameter configurations, as compared to current state-of-the-art results.}
  \label{tab:classification_results}
  \vspace{-9pt}
\end{table}

We performed the optimization on a cluster of Intel$\textsuperscript{\textregistered}$ Xeon Phi$\textsuperscript{\texttrademark}$ coprocessors, with 40 jobs running in parallel using a kernel library that has been highly optimized for efficient computation on the Intel$\textsuperscript{\textregistered}$ Xeon Phi$\textsuperscript{\texttrademark}$ coprocessor\footnote{Available at \url{https://github.com/orippel/micmat}}. For the optimal hyperparameter configuration found, we ran a final experiment for 350 epochs on the entire training set, and report its result.

Our optimal models for CIFAR-10 and CIFAR-100 achieved test errors of $6.37\%$ and $27.4\%$ respectively. A comparison to published state-of-the-art results \cite{GoodfellowWMCB13, WanZZLF13, LinCY13, Lee2014,DBLP:journals/corr/SpringenbergDBR14} is given in Table~\ref{tab:classification_results}.  We see that the parallelized automated hyperparameter tuning procedure obtains models that are highly competitive with the state-of-the-art in just a few sequential steps.

A comprehensive overview of the setup, the architecture, the tuning and the optimum configuration can be found in the supplementary material.

\section{Conclusion}
In this paper, we introduced deep networks for global optimization, or DNGO, which enables efficient optimization of noisy, expensive black-box functions. While this model maintains desirable properties of the GP such as tractability and principled management of uncertainty, it greatly improves its scalability from cubic to linear as a function of the number of observations. We demonstrate that while this model allows efficient computation, its performance is nevertheless competitive with existing state-of-the-art approaches for Bayesian optimization.  We demonstrate empirically that it is especially well suited to massively parallel hyperparameter optimization.

While adaptive basis regression with neural networks provides one approach to the enhancement of scalability, other models may also present promise. One promising line of work, for example by \citet{nicksonautomated}, is to introduce a similar methodology by instead employing the sparse Gaussian process as the underlying probabilistic model~\cite{snelson2005sparse,titsias2009variational,hensman2013bigdatagp}.

\section{Acknowledgements}\small
This work was supported by the Director, Office of Science, Office of Advanced Scientific Computing Research, Applied Mathematics program of the U.S. Department of Energy under Contract No. DE-AC02-05CH11231. This work used resources of the National Energy Research Scientific Computing Center, which is supported by the Office of Science of the U.S. Department of Energy under Contract No. DE-AC02-05CH11231. We would like to acknowledge the NERSC systems staff, in particular Helen He and Harvey Wasserman, for providing us with generous access to the Babbage Xeon Phi testbed.

The image caption generation computations in this paper were run on the Odyssey cluster supported by the FAS Division of Science, Research Computing Group at Harvard University.  We would like to acknowledge the FASRC staff and in particular James Cuff for providing generous access to Odyssey.

Jasper Snoek is a fellow in the Harvard Center for Research on Computation and Society. Kevin Swersky is the recipient of an Ontario Graduate Scholarship (OGS). This work was partially funded by NSF IIS-1421780, the Natural Sciences and Engineering Research Council of Canada (NSERC) and the Canadian Institute for Advanced Research (CIFAR).

\small
\bibliographystyle{icml2015}
\bibliography{draft}

\section*{Supplementary Material}
\appendix
\section{Convolutional neural network experiment specifications}
In this section we elaborate on the details of the network architecture, training and the meta-optimization. In the following subsections we elaborate on the hyperparametrization scheme. The priors on the hyperparameters as well as their optimal configurations for the two datasets can be found in Table~\ref{tab:priors}.

\subsection{Architecture}
The model architecture is specified in Table~\ref{tab:architecture}.  

\begin{table}[ht]
  \centering
  \small{
  \begin{tabular}[b]{lrrr}  
    \toprule
    {\bf Layer type} & \# {\bf Filters} & {\bf Window} & {\bf Stride}\\%
    \cmidrule(r){1-4}%
    Convolution & $96$ & $3\times 3$ & \\
    \cmidrule(r){1-4}%
    Convolution & $96$ & $3\times 3$ & \\
    \cmidrule(r){1-4}%
    Max pooling & & $3\times 3$ & 2 \\
    \cmidrule(r){1-4}%
    Convolution & $192$ & $3\times 3$ & \\
    \cmidrule(r){1-4}%
    Convolution & $192$ & $3\times 3$ & \\
    \cmidrule(r){1-4}%
    Convolution & $192$ & $3\times 3$ & \\
    \cmidrule(r){1-4}%
    Max pooling & & $3\times 3$ & 2 \\
    \cmidrule(r){1-4}%
    Convolution & $192$ & $3\times 3$ & \\
    \cmidrule(r){1-4}%
    Convolution & $192$ & $1\times 1$ & \\
    \cmidrule(r){1-4}%
    Convolution & $10/100$ & $1\times 1$ & \\
    \cmidrule(r){1-4}%
    Global averaging & & $6\times 6$ & \\
    \cmidrule(r){1-4}%
    Softmax & & & \\
   \bottomrule%

  \end{tabular}}
  \captionsetup[table]{position=bottom}
  \captionof{table}{Our convolutional neural network architecture. This choice was chosen to be maximally generic. Each convolution layer is followed by a ReLU nonlinearity.}
  \label{tab:architecture}
\end{table}

\subsection{Data augmentation}
We corrupt each input in a number of ways. Below we describe our parametrization of these corruptions.

\paragraph{HSV} We shift the hue, saturation and value fields of each input by global constants ${b_H\sim U(-B_H, B_H)}$, ${b_S\sim U(-B_S, B_S)}$, ${b_V\sim U(-B_V, B_V)}$. Similarly, we globally stretch the saturation and value fields by global constants ${a_S\sim U(\frac{1}{1+A_S}, 1+A_S)}$, ${a_V\sim U(\frac{1}{1+A_V}, 1+A_V)}$.

\paragraph{Scalings} Each input is scaled by some factor ${s\sim U(\frac{1}{1+S}, 1+S)}$.

\paragraph{Translations} We crop each input to size $27\times 27$, where the window is chosen randomly and uniformly.

\paragraph{Horizontal reflections} Each input is reflected horizontally with a probability of 0.5.

\paragraph{Pixel dropout} Each input element is dropped independently and identically with some random probability $D_0$.

\subsection{Initialization and training procedure}
We initialize the weights of each convolution layer $m$ with i.i.d zero-mean Gaussians with standard deviation $\frac{\sigma}{\sqrt{F_m}}$ where $F_m$ is the number of parameters per filter for that layer. We chose this parametrization to produce activations whose variances are invariant to filter dimensionality. We use the same standard deviation for all layers but the input, for which we dedicate its own hyperparameter $\sigma_I$ as it oftentimes varies in scale from deeper layers in the network.

We train the model using the standard stochastic gradient descent and momentum optimizer. We use minibatch size of 128, and tune the momentum and learning rate, which we parametrize as $1-0.1^M$ and $0.1^L$ respectively. We anneal the learning rate by a factor of $0.1$ at epochs $130$ and $190$. We terminate the training after $200$ epochs.

We regularize the weights of all layers with weight decay coefficient $W$. We apply dropout on the outputs of the max pooling layers, and tune these rates $D_1, D_2$ separately.

\subsection{Testing procedure}
We evaluate the performance of the learned model by averaging its log-probability predictions on 100 samples drawn from the input corruption distribution, with masks drawn from the unit dropout distribution.

\section{Additional figures for image caption generation}
\begin{figure*}[t]
\centering%
\subfigure[\small Heatmap]{
  \includegraphics[width=0.41\linewidth,trim=1cm -1cm 1cm 1cm]{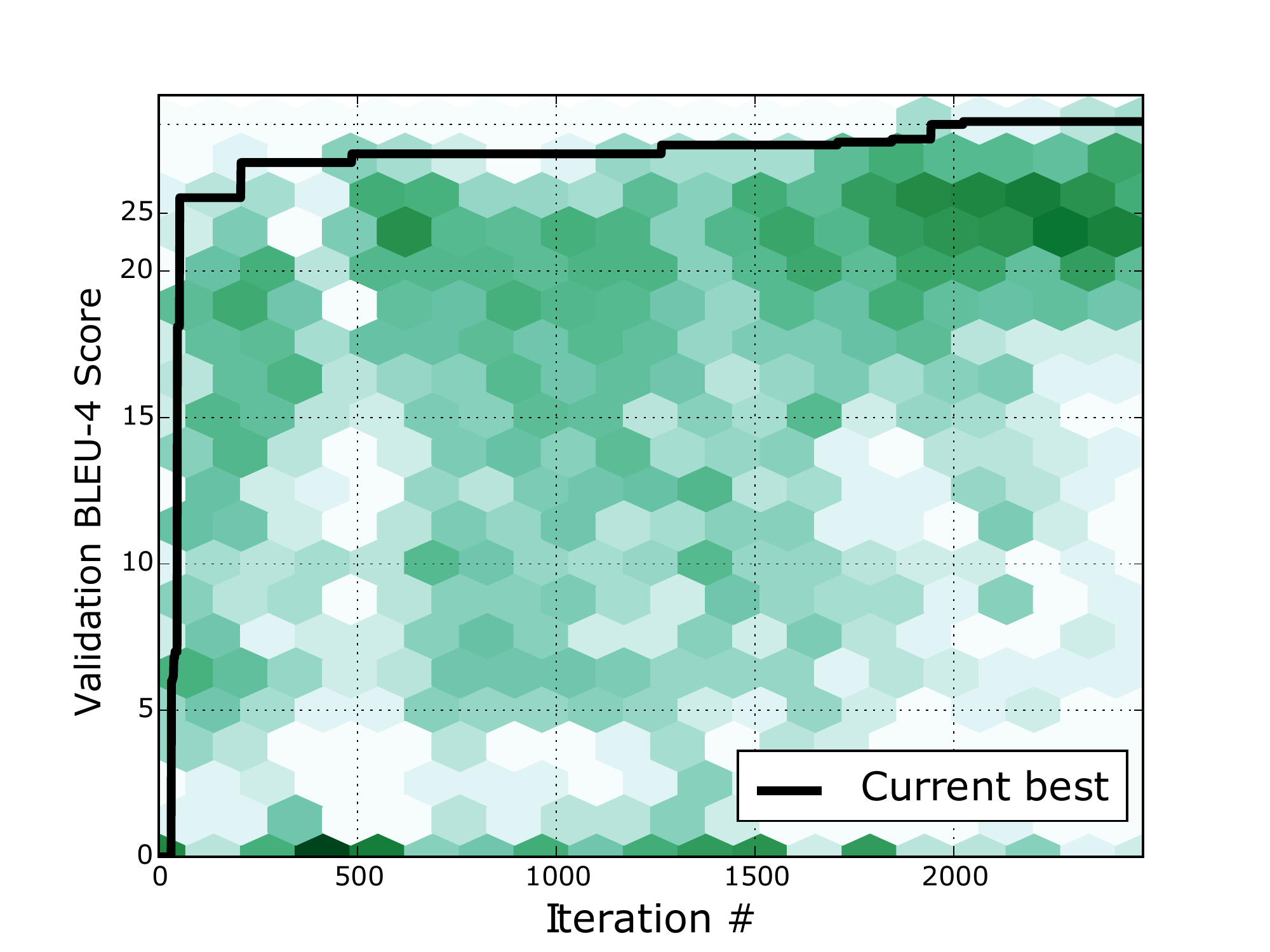}
  \label{fig:heatmap}}%
\subfigure[\small Scatter Plot]{
  \includegraphics[width=0.59\linewidth,trim=1cm 1cm 1cm 1cm]{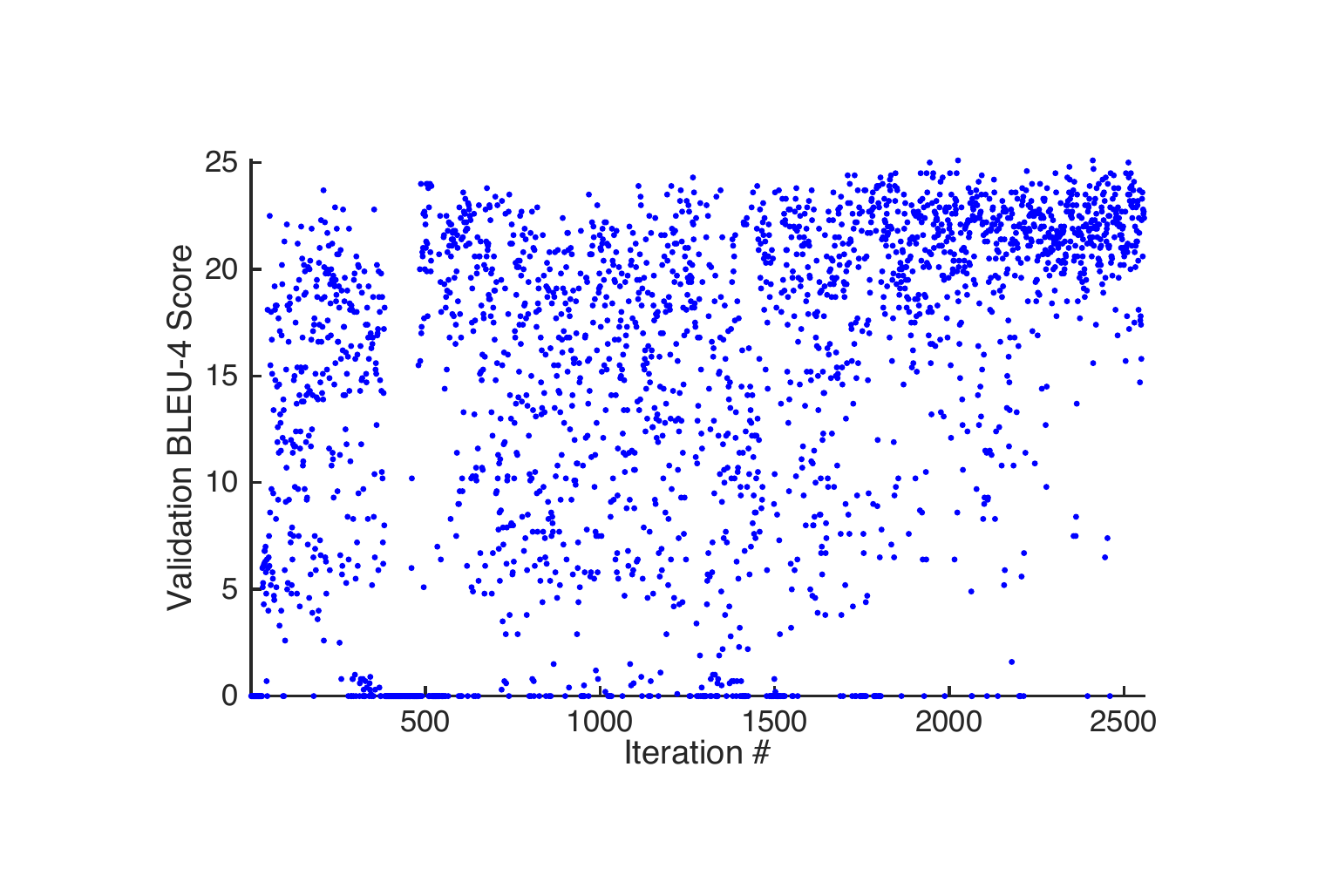}
  \label{fig:scatter}}%

\caption{Validation BLEU-4 Score on MS COCO corresponding to different hyperparameter configurations as evaluated over time. In Figure~\ref{fig:heatmap}, these are represented as a planar histogram, where the shade of each bin indicates the total count within it. The current best validation score discovered is traced in black. Figure~\ref{fig:scatter} shows a scatter plot of the validation score of all the experiments in the order in which they finished.  This projection demonstrates the exploration-versus-exploitation paradigm of Bayesian Optimization, in which the algorithm trades off visiting unexplored parts of the space, and focusing on parts which show promise.
}
\end{figure*}
In Figures~\ref{fig:heatmap} and \ref{fig:scatter} we provide some additional visualization of the results from the image caption generation experiments from Section 4.2 to highlight the behavior of the Bayesian optimization routine.  Both figures show the validation BLEU-4 Score on MS COCO corresponding to different hyperparameter configurations as evaluated over iterations of the optimization procedure. In Figure~\ref{fig:heatmap}, these are represented as a planar histogram, where the shade of each bin indicates the total count within it. The current best validation score discovered is traced in black. Figure~\ref{fig:scatter} shows a scatter plot of the validation score of all the experiments in the order in which they finished.  Validation scores of 0 correspond to constraint violations.  These figures demonstrate the exploration-versus-exploitation paradigm of Bayesian Optimization, in which the algorithm trades off visiting unexplored parts of the space, and focusing on parts which show promise.

\section{Multimodal neural language model hyperparameters}
\subsection{Description of the hyperparameters}
We optimize a total of 11 hyperparameters of the log-bilinear model (LBL). Below we explain what these hyperparameters refer to.

\paragraph{Model} The LBL model has two variants, an additive model where the image features are incorporated via an additive bias term, and a multiplicative that uses a factored weight tensor to control the interaction between modalities.

\paragraph{Context size} The goal of the LBL is to predict the next word given a sequence of words. The context size dictates the number of words in this sequence.

\paragraph{Learning rate, momentum, batch size} These are optimization parameters used during stochastic gradient learning of the LBL model parameters. The optimization over learning rate
is carried out in log-space, but the proposed learning rate is exponentiated before being passed
to the training procedure.

\paragraph{Hidden layer size} This controls the size of the joint hidden representation for words and images.

\paragraph{Embedding size} Words are represented by feature embeddings rather than one-hot vectors. This is the dimensionality of the embedding.

\paragraph{Dropout} A regularization parameter that determines the amount of dropout to be added to the hidden layer.

\paragraph{Context decay, Word decay} $\mathcal{L}_2$ regularization on the input and output weights respectively. Like the learning rate, these are optimized in log-space as they vary over several orders of magnitude.

\paragraph{Factors} The rank of the weight tensor. Only relevant for the multiplicative model.

\begin{table*}[t]
  \centering
  \begin{tabular}[b]{lllll}
    \toprule
    {\bf Hyperparameter} & {\bf Notation} & {\bf Support of prior} & {\bf CIFAR-10 Optimum} & {\bf CIFAR-100 Optimum}\\%
    \cmidrule(r){1-5}%
    Momentum & $M$ & $[0.5, 2]$ & $1.6242$ & $1.3339$ \\
    Learning rate & $L$ & $[1, 4]$ & $2.7773$ & $2.1205$ \\
    Initialization deviation & $\sigma_I$ & $[0.5, 1.5]$ & $0.83359$ & $1.5570$ \\
    Input initialization deviation & $\sigma$ & $[0.01, 1]$ & $0.025370$ & $0.13556$ \\
    Hue shift & $B_H$ & $[0, 45]$ & $31.992$ & $19.282$ \\
    Saturation scale & $A_S$ & $[0, 0.5]$ & $0.31640$ & $0.30780$ \\
    Saturation shift & $B_S$ & $[0, 0.5]$ & $0.10546$ & $0.14695$ \\
    Value scale & $A_S$ & $[0, 0.5]$ & $0.13671$ & $0.13668$ \\
    Value shift & $B_S$ & $[0, 0.5]$ & $0.24140$ & $0.010960$ \\
    Pixel dropout & $D_0$ & $[0, 0.3]$ & $0.19921$ & $0.00056598$ \\
    Scaling & $S$ & $[0, 0.3]$ & $0.24140$ & $0.12463$ \\
    L2 weight decay & $W$ & $[2, 5]$ & $4.2734$ & $3.1133$ \\
    Dropout 1 & $D_1$ & $[0, 0.7]$ & $0.082031$ & $0.081494$ \\
    Dropout 2 & $D_2$ & $[0, 0.7]$ & $0.67265$ & $0.38364$ \\
   \bottomrule%
\end{tabular}
  \captionsetup[table]{position=bottom}
  \captionof{table}{Specification of the hyperparametrization scheme, and optimal hyperparameter configurations found.}
  \label{tab:priors}
\end{table*}

\begin{table*}[t]
  \centering
  \begin{tabular}{llll}
    \toprule
    {\bf Hyperparameter} & {\bf Support of prior} & {\bf Notes} & {\bf COCO Optimum}\\%
    \cmidrule(r){1-4}%
    Model & \{additive,multiplicative\} & & additive \\
    Context size & $[3,25]$ & & $5$ \\
    Learning rate & $[0.001,10]$ & Log-space & $0.43193$ \\
    Momentum & $[0,0.9]$ & & $0.23269$ \\
    Batch size & $[20,200]$ & & $40$ \\
    Hidden layer size & $[100,2000]$ & & $441$ \\
    Embedding size & $\{50,100,200\}$ & & $100$ \\
    Dropout & $[0,0.7]$ & & $0.14847$ \\
    Word decay & $[10^{-9},10^{-3}]$ & Log-space & $2.98456^{-7}$ \\
    Context decay & $[10^{-9},10^{-3}]$ & Log-space & $1.09181^{-8}$ \\
    Factors & $[50,200]$ & Multiplicative model only & - \\
   \bottomrule%
\end{tabular}
  \captionsetup[table]{position=bottom}
  \captionof{table}{Specification of the hyperparametrization scheme, and optimal hyperparameter configurations found for the multimodal neural language model. For parameters marked log-space, the log is given to the Bayesian optimization routine and the result is exponentiated before being passed into the multimodal neural language model for training. Square brackets denote a range of parameters, while curly braces denote a set of options.}
  \label{tab:priors2}
\end{table*}

\end{document}